\documentclass{llncs}
\usepackage{makeidx}  
\usepackage{graphicx}
\usepackage{wrapfig}
\usepackage{subfig}
\usepackage{amsmath}
\usepackage{multirow}
\usepackage{array}
\usepackage{bbm}

\begin{document}
\mainmatter              
\title{Full Quantification of Left Ventricle via Deep Multitask Learning Network Respecting \\Intra- and Inter-Task Relatedness}

\titlerunning{Full Quantification of Left Ventricle}  
\author{Wufeng Xue \and Andrea Lum \and Ashley Mercado \and Mark Landis \and James Warrington \and Shuo Li*}
\authorrunning{W. Xue, A. Lum, A. Mercado, et al.} 
\institute{Department of Medical Imaging, Western University, ON, Canada\\
Digital Imaging Group of London, ON, Canada\\
\email{slishuo@gmail.com}\\ 
}

\maketitle              

\begin{abstract}
Cardiac left ventricle (LV) quantification is among the most clinically important tasks for identification and diagnosis of cardiac diseases, yet still a challenge due to the high variability of cardiac structure and the complexity of temporal dynamics. Full quantification, i.e., to simultaneously quantify all LV indices including two areas (cavity and myocardium), six regional wall thicknesses (RWT), three LV dimensions, and one cardiac phase, is even more challenging since the uncertain relatedness intra and inter each type of indices may hinder the learning procedure from better convergence and generalization. 
In this paper, we propose a newly-designed multitask learning network (FullLVNet), which is constituted by a deep convolution neural network (CNN) for expressive feature embedding of cardiac structure; two followed parallel recurrent neural network (RNN) modules for temporal dynamic modeling; and four linear models for the final estimation. During the final estimation, both intra- and inter-task relatedness are modeled to enforce improvement of generalization: 1) respecting intra-task relatedness, group lasso is applied to each of the regression tasks for sparse and common feature selection and consistent prediction; 2) respecting inter-task relatedness, three phase-guided constraints are proposed to penalize violation of the temporal behavior of the obtained LV indices. Experiments on MR sequences of 145 subjects show that FullLVNet achieves high accurate prediction with our intra- and inter-task relatedness, leading to MAE of 190mm$^2$, 1.41mm, 2.68mm for average areas, RWT, dimensions and error rate of 10.4\% for the phase classification. This endows our method a great potential in comprehensive clinical assessment of global, regional and dynamic cardiac function.

\keywords{left ventricle quantification, recurrent neural network, multi-task learning, task relatedness}
\end{abstract}

\section{Introduction}

\begin{figure}[t]
	\centering
	\includegraphics[width=10cm]{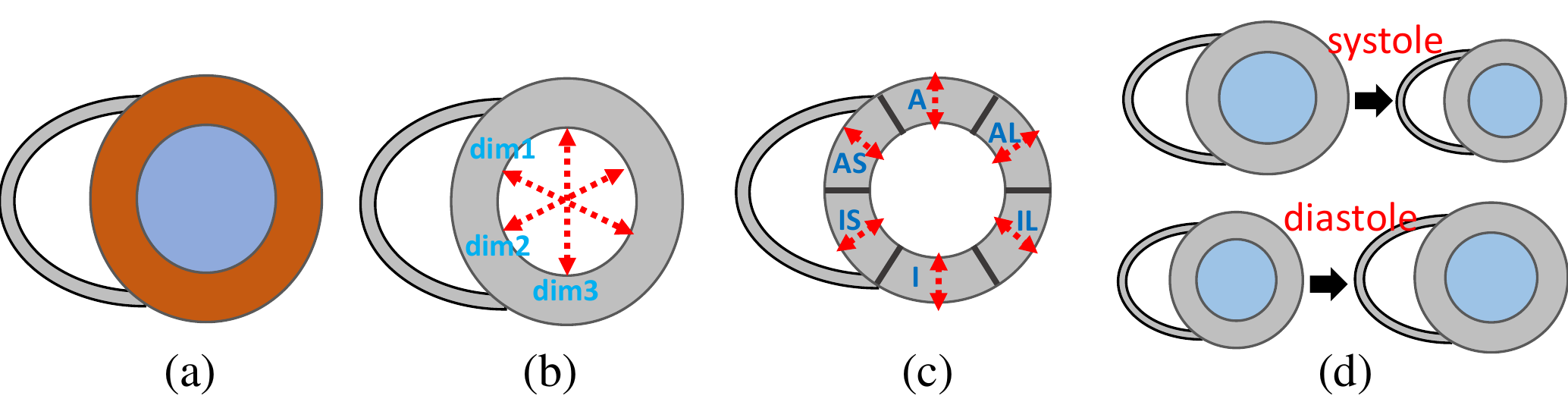}
	\caption{Illustration of LV indices to be quantified for short-axis view cardiac image. (a) Cavity (blue) and myocardium (orange) areas. (b) directional dimensions of cavity (red arrows). (c) Regional wall thicknesses (red arrows). A: anterior; AS: anterospetal; IS: inferoseptal; I: inferior;  IL: inferolateral; AL: anterolateral. (d) Phase (systole or diastole).}
	\label{fig_indices}
\end{figure}

Quantification of left ventricle (LV) from cardiac imaging is among the most clinically important and most frequently demanded tasks for identification and diagnosis of cardiac disease~\cite{karamitsos2009role}, yet still a challenging task due to the high variability of cardiac structure across subjects and the complicated global/regional temporal dynamics. 
Full quantification, i.e., to simultaneously quantify all LV indices including two areas, six regional wall thicknesses (RWT), three LV dimension, and one phase (as shown in Fig.~\ref{fig_indices}), providing more detailed information for comprehensive cardiac function assessment, is even more challenging since the uncertain relatedness intra and inter each type of indices may hinder the learning procedure from better convergence and generalization. In this work, we propose a newly-designed deep multitask learning network FullLVNet for full quantification of LV respecting both intra- and inter-task relatedness.

In clinical practice, obtaining reliable quantification is subjected to measuring on segmented myocardium, which is usually obtained by manually contouring the borders of myocardium~\cite{suinesiaputra2015quantification} or manual correction of contours~\cite{attili2010quantification,kawel2015normal} generated by LV segmentation algorithms~\cite{peng2016review}. However, manually contouring is time-consuming, of high inter-observer variability, and typically limited to the end diastolic (ED) and end systolic (ES) frames, which makes it insufficient for dynamic function analysis. LV segmentation algorithms, despite the recent advances, is still a difficult problem due to the lack of edge information and presence of shape variability. Most existing segmentation methods for cardiac MR images~\cite{peng2016review,petitjean2011review,ayed2012max} requires strong prior information and user interaction to obtain reliable results, which may prevent them from efficient clinical application. 

In recent years, direct methods without segmentation have grown in popularity in cardiac volumes estimation~\cite{afshin2012global,afshin2014regional,wang2014direct,zhen2015direct,zhen2014direct,zhen2015multi,zhen2017direct}.
Although these methods obtained effective performance by leveraging state-of-art machine learning techniques, they suffer from the following limitations. 
1) Lack of powerful task-aware representation. The vulnerable hand-crafted or task-unaware features are not capable of capturing sufficient task-relevant cardiac structures. 
2) Lack of temporal modeling. Independently handling each frame without assistance from neighbors can not guarantee the consistency and accuracy. 
3) Not end-to-end learning. The separately learned representation and regression models cannot be optimal for each other.
4) Not full quantification. Only cardiac volume alone is not sufficient for comprehensive global, regional and dynamic function assessment. 

In this paper, we propose a newly-designed multitask learning network (FullLVNet), which is constituted by a specially tailored deep CNN for expressive feature embedding; two followed parallel RNN modules for temporal dynamic modeling; and four linear models for the final estimation. During the final estimation, FullLVNet is capable of improving the generalization by 1) modeling intra-task relatedness through group lasso regularization within each regression task; and 2) modeling inter-task relatedness with three phase-guided constraints that penalize violation of the temporal behavior of LV indices. After being trained with a two-step strategy, FullLVNet is capable of delivering accurate results for all the considered indices of cardiac LV.

\section{Multitask learning for full quantification of cardiac LV}

The proposed FullLVNet models full quantification of cardiac LV as a multitask learning problem. Three regression tasks $\{y_{area}^{s,f}, y_{dim}^{s,f}, y_{rwt}^{s,f}\}$ and one classification task $y_{phase}^{s,f}$ are simultaneously learned to predict frame-wise values of the above mentioned LV indices from cardiac MR sequences $\mathcal{X}=\{X^{s,f}\}$, where $s=1\cdots S$ indexes the subject and $f=1\cdots F$ indexes the frame. 
The objective of FullLVNet is:
\begin{equation}\label{eq_objective}
W_{optimal}=\min_{W}\frac{1}{S\times F}\sum_{s,f}\sum_{t}L_t(\hat{y}_t^{s,f}(X^{s,f}|W),y_t^{s,f})+\lambda \mathcal{R}(W)
\end{equation}
where $t\in \{area, dim, rwt, phase\}$ denotes a specific task, $\hat{y}_t$ is the estimated results for task $t$, $L_t$ is the loss function of task $t$ and $\mathcal{R}(W)$ denotes regularization of parameters in the network.

\begin{figure}[h]
	\centering
	\includegraphics[width=10cm]{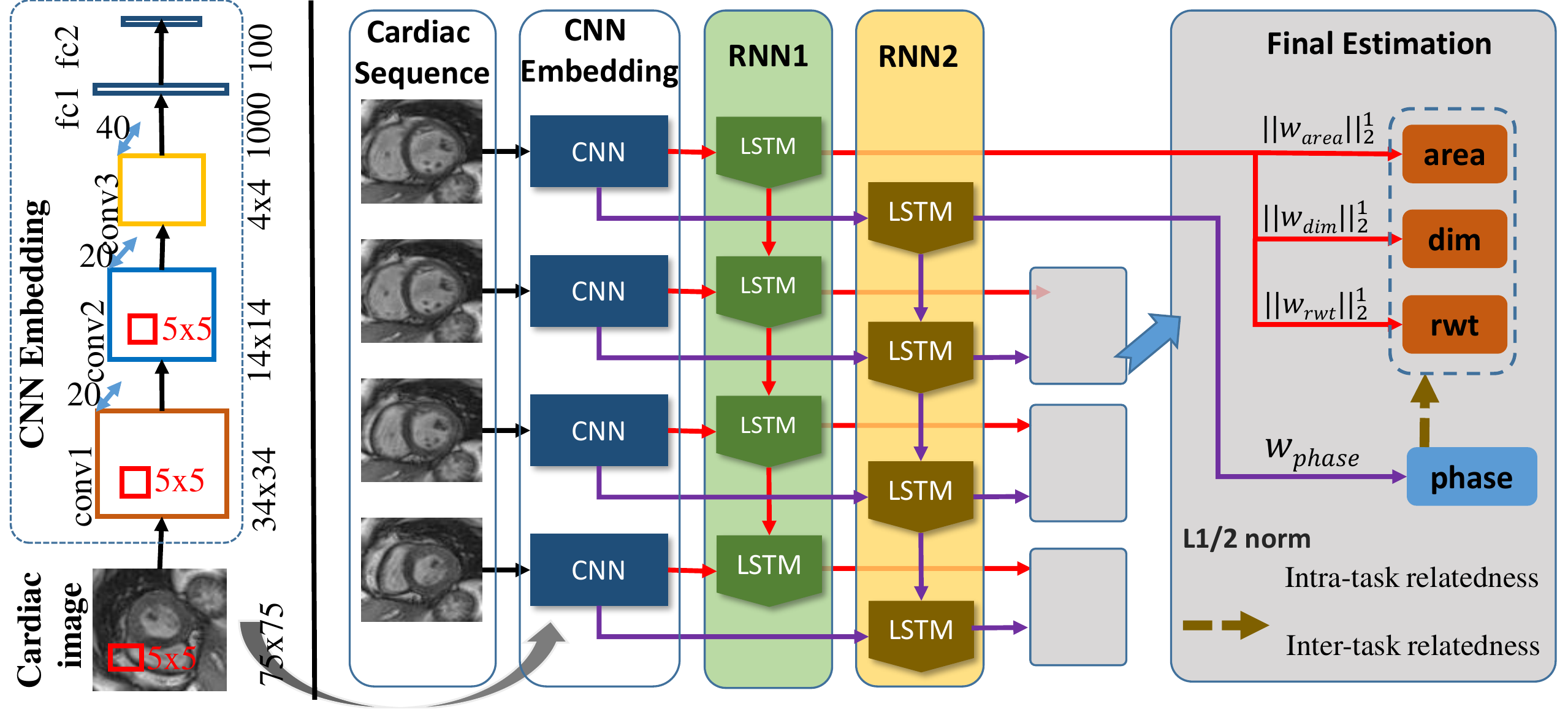}
	\caption{Overview of FullLVNet, which combines a deep CNN network (details shown in the left) for feature embedding, two RNN modules for temporal dynamic modeling, and four linear models for final estimation. Intra- and inter-task relatedness are modeled in the final estimation to improve generalization.}
	\label{fig_network}
\end{figure}

\subsection{Architectures of FullLVNet}
Fig.~\ref{fig_network} shows the overview of FullLVNet. A deep CNN is firstly designed to extract from cardiac images expressive and task-aware feature, which is then fed to the RNN modules for temporal dynamic modeling. Final estimations are given by four linear models with the output of RNN modules as input. To improve generalization of FullLVNet, both intra- and inter-task relatednesses are carefully modeled through group lasso and phase-guided constraints for the linear models.

\textbf{CNN for deep feature embedding}
To obtain expressive and task-aware features, we design a specially tailored deep CNN for cardiac images, as shown in the left of Fig.~\ref{fig_network}. 
Powerful representations can be obtained by transfer learning~\cite{shin2016deep} from well-known deep architectures in computer vision for applications with limited labeled data. However, transfer learning may incur 1) restriction of network architecture, resulting in incompatible or redundant model; and 2) restriction of input channel and dimension, leading to requirement of image resizing and channel expanding. 
We reduce the number of filters for each layer to avoid model redundancy. As for the kernel size of convolution and pooling, $5\times 5$, instead of the frequently used $3\times 3$, is deployed to introduce more shift invariance. Dropout and batch normalization are adopted to alleviate the training procedure. As can be seen in our experiments, our CNN is very effective for cardiac images even without transfer learning. As a feature embedding network, our CNN maps each cardiac image $X^{s,f}$ into a fixed-length low dimension vector $e^{s,f}=f_{cnn}(X^{s,f}|w_{cnn})\in \mathcal{R}^{100}$. 

\textbf{RNNs for temporal dynamic modeling}
Accurate modeling of cardiac temporal dynamic assistants the quantification of current frame with information from neighbors. RNN, especially when LSTM  units~\cite{graves2012supervised} are deployed, is specialized in temporal dynamic modeling and has been employed in cardiac image segmentation~\cite{poudel2016recurrent} and key frame recognition~\cite{kong2016recognizing} in cardiac sequences.
In this work, two RNN modules, as shown by the green and yellow blocks in Fig.~\ref{fig_network}, are deployed for the regression tasks and the classification task. For the three regression tasks, the indices to be estimated are mainly related to the spatial structure of cardiac LV in each frame. For the classification task, the cardiac phase is mainly related to the structure difference between successive frames. Therefore, the two RNN modules are designed to capture these two kinds of dependencies. The outputs of RNN modules are $\{h_m^{s,1},,,,h_m^{s,F}\}=f_{rnn}([e^{s,1},...e^{s,F}]|w_{m}),m\in\{rnn1,rnn2\}$.

\textbf{Final estimation} 
With the outputs of RNN modules, all the LV indices can be estimated with a linear regression/classification model:  
\begin{equation}\label{eq_linear_model}
\begin{cases}
\hat{y}_{t}^{s,f}=w_{t}h_{rnn1}^{s,f}+b_t, ~where ~t\in\{area, dim, rwt\} \\
p(\hat{y}_{t}^{s,f}=0)=\frac{1}{1+\exp(w_{t}h_{rnn2}^{s,f}+b_t)}, ~t=phase
\end{cases}
\end{equation}
where $w_{t}$ and $b_t$ are the weight and bias term of the linear model for task $t$, $0$ and $1$ denote the two cardiac phase $Diastole$ and $Systole$, and $p(\hat{y}_{phase}^{s,f}=1)=1-p(\hat{y}_{phase}^{s,f}=0)$.  
For the loss function in~(\ref{eq_objective}), Euclidean distance and cross-entropy are employed for the regression tasks and the classification task, respectively.

\begin{equation}
L_{t} = \begin{cases}
\frac{1}{2}\|\hat{y}_t^{s,f}-y_t^{s,f}\|_2^2, ~where ~t\in\{area, dim, rwt\}\\
-\log p(\hat{y}_{t}^{s,f}=y_{t}^{s,f}),~t=phase
\end{cases}
\end{equation}

\subsection{Intra-task and inter-task relatedness}
Significant correlations exist between the multiple outputs of each task and those of different tasks, and are referred as intra- and inter-task relatedness. Intra-task relatedness can be effectively modeled by the well-known group lasso regularization, while inter-task relatedness is modeled by three phase-guided constraints. Improved generalization can be achieved with both of them fully leveraged in our FullLVNet.  

\textbf{Intra-task relatedness based on group lasso}
Group lasso, also known as L1/L2 regularization, can perfectly model relatedness within groups of outputs, i.e, the three regression tasks. It enforces common feature selection cross related outputs with the L2 norm, and encourages sparse selection of the most related features with the L1 norm for each task. In this way, the relevant features of different tasks can be well disentangled. To leverage this advantage, group lasso is applied to the weight parameters of the three regression models in~(\ref{eq_linear_model}). 
\begin{equation}
\mathcal{R}_{intra}=\sum_t \sum_i\|w_t(i)\|_2, ~for ~t\in\{area, dim, rwt\}
\end{equation}
where $w_t(i)$ denotes the $i$th column of $w_t$. 
 
\textbf{Inter-task relatedness based on phase-guided constraints}
Three phase-guided constraints are proposed to model inter-task relatedness, i.e, the cardiac phase and other LV indices. Cardiac phase indicates the temporal dynamics of LV myocardium in a cardiac cycle. Other LV indices change accordingly with cardiac phase: 1) cavity area and LV dimensions increase in the diastole phase and decrease in the systole phase; 2) myocardium area and RWT decrease in the diastole phase and increase in the systole phase. Effectively modeling such an intrinsic phase-guided relatedness would ensure that the estimated LV indices are consistent with the temporal dynamics of LV. 
To penalize violation of these inter-task relatednesses, three phase-guided constraints are applied to the predicted results of areas, dimensions and RWT.
\begin{equation}
\begin{split}
\mathcal{R}_{inter}^{area}=\frac{1}{2S\times F}&\sum_{s,f}[\mathbbm{1}(y_{phase}^{s,f}=0)(\max(-z_{area}^{s,f,1},0)+\max(z_{area}^{s,f,2},0))\\
&+\mathbbm{1}(y_{phase}^{s,f}=1)(\max(z_{area}^{s,f,1},0)+\max(-z_{area}^{s,f,2},0))]
\end{split}
\end{equation}
\begin{equation}
\mathcal{R}_{inter}^{dim}=\frac{1}{S\times F}\sum_{s,f}[\mathbbm{1}(y_{phase}^{s,f}=0)\max(-\bar{z}_{dim}^{s,f},0)+ \mathbbm{1}(y_{phase}^{s,f}=1)\max(\bar{z}_{dim}^{s,f},0)]
\end{equation}
\begin{equation}
\mathcal{R}_{inter}^{rwt}=\frac{1}{S\times F}\sum_{s,f}[\mathbbm{1}(y_{phase}^{s,f}=0)\max(\bar{z}_{rwt}^{s,f},0)+ \mathbbm{1}(y_{phase}^{s,f}=1)\max(-\bar{z}_{rwt}^{s,f},0)]
\end{equation}
where $\mathbbm{1}(\cdot)$ is the indicator function, $z_t^{s,f}=\hat{y}_t^{s,f}-\hat{y}_t^{s,f-1}, for~t\in\{area,dim,rwt\}$, $z_t^{s,f,i}$ denotes the $i$th output of $z_t$ and $\bar{z}_t$ denotes the average value of $z_t$ across its multiple outputs. Totally, our regularization term becomes
\begin{equation}
\mathcal{R}(W)=\lambda_1\mathcal{R}_{intra}+\lambda_2(\mathcal{R}_{inter}^{area}+\mathcal{R}_{inter}^{dim}+\mathcal{R}_{inter}^{rwt})
\end{equation}

\section{Dataset and Configurations}

Our FullLVNet is validated with short-axis cardiac MR images of 145 subjects. Temporal resolution is 20 frames per cardiac cycle, resulting in a total of 2900 images in the dataset. The pixel spacings range from 0.6836 mm/pixel to 2.0833 mm/pixel, with mode of 1.5625 mm/pixel. The ground truth values are computed from manually obtained contours of LV myocardium. Within each subject, frames are labeled as either Diastole phase or Systole phase, according to the obtained values of cavity area.
In our experiments, two landmarks, i.e, junctions of the right ventricular wall with the left ventricular, are manually marked for each image to provide reference for ROI cropping and the LV myocardial segments division. The cropped images are resized to $80\times80$. The network is implemented by Caffe with SGD solver. Five-fold cross validation is employed for performance evaluation and comparison. Data augmentation is conducted by randomly cropping images of size $75\times 75$ from the resized image. 

\textbf{Two-step strategy training} 
We apply a two-step strategy for training our network to alleviate the difficulties caused by the different learning rate and loss function in multitask learning~\cite{zhang2010convex,zhang2014facial}. 
Firstly the CNN embedding, the first RNN module and the three regression models are learned together with no back propagation from the classification task, to obtain accuracy prediction for the regression tasks; with the obtained CNN embedding, the second RNN module and the linear classification model are then learned while the rest of the network are kept frozen. As shown in the experiments, such a strategy delivers excellent performance for all the considered tasks.

\section{Results and Analysis}
FullLVNet is extensively validated under different configurations in Table.~\ref{table_results}. From the last column, we can draw that FullLVNet successfully delivers accurate predictions for all the considered indices, with average Mean Absolute Error (MAE) of 1.41$\pm0.72mm$, 2.68$\pm1.64mm$, 190$\pm128mm^2$ for RWT, dimension, and areas. For reference, the maximums of these indices in our dataset are 24.4$mm$, 81.0$mm$, 4936$mm^2$. Error rate (1-accuracy) for phase identification is 10.4\%.
Besides, the effectivenesses of intra- and inter-task relatedness are also demonstrated by the results in the third and fourth column: intra-task relatedness brings clearly improvements for all the tasks, while inter-task relatedness further brings moderate improvement. Compared to the recent direct multi-feature based method~\cite{zhen2015multi}, which we adapt to our full quantification task, FullLVNet shows remarkable advantages even without intra- and inter-task relatedness. 

\begin{table*}[t]
	\caption{Performance of FullLVNet under different configurations (e.g, intra/N means only intra-task relatedness is included) and its competitor for LV quantification. Mean Absolute Error (MAE) is used for the three regression tasks and prediction error rate is used for the phase identification task.}
	\label{table_results}
	\centering
	\begin{tabular}{l|c|ccc}
		\hline
		\multirow{2}{*}{Method} & Multi-features & \multicolumn{3}{c}{FullLVNet}\\ \cline{3-5}
		&\cite{zhen2015multi} &N/N&intra/N&intra/inter\\
		\hline
		\multicolumn{5}{c}{RWT (mm)}\\
		\hline
		IS&1.70$\pm$1.47&1.42$\pm$1.21&1.39$\pm$1.10&\textbf{1.32$\pm$1.09}\\
		I&1.71$\pm$1.34&1.53$\pm$1.25&1.48$\pm$1.16&\textbf{1.38$\pm$1.10}\\
		IL&1.97$\pm$1.54&1.74$\pm$1.43&1.61$\pm$1.29&\textbf{1.57$\pm$1.35}\\
		AL&1.82$\pm$1.41&1.59$\pm$1.31&\textbf{1.53$\pm$1.06}&1.60$\pm$1.36\\
		A&1.55$\pm$1.33&1.36$\pm$1.17&\textbf{1.32$\pm$1.06}&1.34$\pm$1.11\\
		AS&1.68$\pm$1.43&1.43$\pm$1.24&1.37$\pm$1.10&\textbf{1.26$\pm$1.10}\\
		\hline
		Average&1.73$\pm$0.97&1.51$\pm$0.81&1.45$\pm$0.69&\textbf{1.41$\pm$0.72}\\
		\hline
		\multicolumn{5}{c}{Dimension (mm)}\\
		\hline
         dim1&3.53$\pm$2.77&2.87$\pm$2.23&2.69$\pm$2.05&\textbf{2.62$\pm$2.09}\\
         dim2&3.49$\pm$2.87&2.96$\pm$2.35&2.67$\pm$2.15&\textbf{2.64$\pm$2.12}\\
         dim3&3.91$\pm$3.23&2.92$\pm$2.48&2.70$\pm$2.22&\textbf{2.77$\pm$2.22}\\
         \hline
         Average&3.64$\pm$2.61&2.92$\pm$1.89&2.69$\pm$1.67&\textbf{2.68$\pm$1.64}\\
		\hline
		\multicolumn{5}{c}{Area (mm$^2$)}\\
		\hline
		cavity&231$\pm$193&205$\pm$182&182$\pm$152&\textbf{181$\pm$155}\\
		myocardium&291$\pm$246&204$\pm$195&205$\pm$168&\textbf{199$\pm$174}\\
		\hline
		Average&261$\pm$165&205$\pm$145&194$\pm$125&\textbf{190$\pm$128}\\
		\hline
		\multicolumn{5}{c}{Phase (\%)}\\
		\hline
		phase&22.2&13.0&11.4&\textbf{10.4}\\
		\hline
	\end{tabular}
\end{table*}

\section{Conclusions}
We propose a multitask learning network FullLVNet for full quantification of LV, which includes three regression tasks and one classification task. By taking advantages of expressive feature embeddings from deep CNN and effective dynamic temporal modeling from RNN, and leveraging intra- and inter-task relatedness with group lasso regularization and phase-guided constraints, FullLVNet is capable of delivering state-of-art accuracy for all the tasks considered.

\bibliographystyle{splncs03}
\bibliography{cardiac}

\end{document}